\documentclass{article}

\PassOptionsToPackage{numbers,compress}{natbib}

\usepackage[eandd,final]{neurips_2026}
\usepackage[accsupp]{axessibility}  
\usepackage[utf8]{inputenc}
\usepackage{graphicx}
\usepackage{subcaption}
\usepackage{wrapfig}
\usepackage{float}
\usepackage{caption}
\usepackage[table]{xcolor}
\usepackage{tikz}

\usepackage{eccvabbrv}

\usepackage{graphicx}
\usepackage{natbib}
\usepackage{booktabs}
\usepackage[utf8]{inputenc}
\usepackage{graphicx}
\usepackage{tabularx}
\usepackage[most]{tcolorbox}

\usepackage{booktabs}

\usepackage[utf8]{inputenc} 
\usepackage[T1]{fontenc}

\usepackage{hyperref}       
\usepackage{url}            
\usepackage{booktabs}       
\usepackage{amsmath,amssymb}
\usepackage{amsfonts}       
\usepackage{nicefrac}       
\usepackage{microtype}      
\usepackage{xcolor}         
\usepackage{placeins}

\title{CurveBench: A Benchmark for Exact Topological Reasoning over Nested Jordan Curves}

\author{%
  Amirreza Mohseni \\
  University of Oxford \\
  \texttt{amirreza.mohseni@cs.ox.ac.uk}
  \And
  Mona Mohammadi \\
  Cornell University \\
  \texttt{mm3325@cornell.edu}
  \AND
  Morteza Saghafian \\
  TU Wien \\
  \texttt{msaghafi@ac.tuwien.ac.at}
  \And
  Naser Talebizadeh Sardari \\
  Pennsylvania State University \\
  \texttt{nzt5208@psu.edu}
}

\hypersetup{
  pdftitle={CurveBench: A Benchmark for Exact Topological Reasoning over Nested Jordan Curves},
  pdfauthor={Amirreza Mohseni, Mona Mohammadi, Morteza Saghafian, Naser Talebizadeh Sardari},
  colorlinks=true, linkcolor=blue!50!black, citecolor=blue!50!black, urlcolor=blue!50!black
}
\begin{document}

\maketitle

\begin{abstract}
We introduce CurveBench, a benchmark for visual topological reasoning. Given an image of disjoint closed curves, a model must recover the tree of adjacent regions rooted at the exterior. The benchmark contains 756 hand-drawn images, ranging from simple configurations to maze-like and densely packed arrangements. Each image has a manually verified reference tree. We evaluate predictions by exact tree matching with the exterior root fixed. Gemini 3.1 Pro, the strongest model in our evaluation, achieves \textbf{71.1\%} exact-match accuracy on CurveBench-Easy and \textbf{28.7\%} on CurveBench-Hard. A Qwen3-VL-8B checkpoint trained using reinforcement learning with verifiable rewards reaches \textbf{31.7\%} on Easy, compared with \textbf{21.7\%} for its Thinking base model at a 16,384-token generation budget. The same comparison yields a gain of 1.1 percentage points on Hard.
\end{abstract}

\section{Introduction}
\label{sec:intro}
Images of disjoint curves arise naturally in many areas of mathematics and the applied sciences. From a topological perspective, families of pairwise disjoint curves encode essential information about connectivity, separation, and the decomposition of the plane into regions. Their arrangement, nesting, and adjacency determine the global structure of the underlying space and often admit a rich combinatorial description.

A classical example appears in topographic maps, where contour lines form disjoint level sets representing elevation and partition the terrain into meaningful regions. More generally, level sets of polynomials and other functions produce structured families of non-intersecting curves whose topology reflects critical points and qualitative features of the function. In biology, similar patterns arise in cellular tissues, anatomical cross sections, and growth structures, where disjoint boundaries organize and constrain spatial form.

At the same time, interpreting such images remains a significant challenge for large language models. Although these models excel at processing text, extracting and reasoning about geometric and topological structure in images, especially when it depends on subtle relations such as disjointness, nesting, and separation, is far from fully understood.

To systematically study topological reasoning from images, we introduce a new dataset, called CurveBench, consisting of synthetic and structured images formed by collections of pairwise disjoint Jordan curves in the plane. Each image induces a well defined nesting structure, where curves enclose regions without intersecting one another. We formulate the core task as extracting this nestedness relation directly from the image, producing a rooted tree in which each node corresponds to a region and each edge represents the presence of a common boundary curve separating two regions. By isolating containment and separation as the primary signal, CurveBench provides a controlled benchmark for evaluating a model’s ability to extract structured topological representations from visual input.

The hierarchical nesting and topological complexity that CurveBench introduces highlight the limitations of current state-of-the-art LLMs in capturing topological structures from images. Classical contour-following methods can assist in extracting containment hierarchies, but their outputs on rasterized images require verification. Our results demonstrate that modern Vision-Language Models (VLMs) still struggle with this basic topological capability. CurveBench serves as a diagnostic baseline to evaluate and close this gap, providing a structured training signal for learning combinatorial relationships that remain difficult for current attention-based visual encoders. Our contributions are:
\begin{itemize}
\item We introduce CurveBench, a controlled benchmark for exact visual topological reasoning over pairwise non-intersecting Jordan curves.
\item We define a deterministic structured prediction task, evaluation protocol, parser, and exact rooted-tree matching metric.
\item We release datasets, Croissant metadata, evaluation environments, ground-truth generation code, and training artifacts to support reproducible evaluation.
\item We benchmark a range of frontier and open-weight VLMs, showing that current models remain far from solving exact containment-tree recovery.
\item We evaluate released RLVR-fine-tuned open-weight VLMs, finding higher accuracy on CurveBench-Easy and limited transfer to CurveBench-Hard.
\end{itemize}

\section{Related work}
\label{sec:rel}

\paragraph{Structured prediction from images.}
A central direction in computer vision is mapping visual input to structured outputs such as trees, graphs, or sequences. Classical approaches connect image boundaries to hierarchical region representations, for example via Ultrametric Contour Maps (UCM), where contours induce nested region trees \cite{arbelaez2011contours}. More recent work directly predicts structured representations from images, including road-network graphs \cite{bastani2018roadtracer} and polygonal or map structures \cite{li2019polymapper}. Scene graph parsing methods further model relational structure over objects \cite{zellers2018neuralmotifs,krishna2017visualgenome}.

In parallel, structured outputs have been reformulated as sequence generation problems. Pix2Seq models object detection as token prediction \cite{chen2022pix2seq}, while Pix2Struct generalizes this paradigm to broader image-to-structure tasks via pretraining \cite{lee2023pix2struct}. Set-based prediction frameworks such as DETR demonstrate that structured outputs can be learned end-to-end without task-specific pipelines \cite{carion2020detr}. 

In contrast to these works, our task predicts a \emph{rooted containment tree} induced by planar regions and requires exact recovery of all parent--child relations, making the problem strictly combinatorial rather than approximate or geometric.

\paragraph{Diagram understanding and visual reasoning.}
CurveBench is closely related to diagram understanding and visual reasoning benchmarks. AI2D and IconQA study reasoning over diagrams through parsing and question answering \cite{kembhavi2016ai2d,lu2021iconqa}. Diagnostic datasets such as CLEVR \cite{johnson2017clevr} and GQA \cite{hudson2019gqa} emphasize compositional reasoning under controlled settings, while spatial reasoning benchmarks such as VSR highlight persistent challenges in modeling fine-grained spatial relations \cite{liu2023vsr}. 

Unlike these benchmarks, which typically require answering queries, CurveBench isolates a single global structural task: reconstructing the full containment hierarchy induced by disjoint curves. This enables deterministic evaluation of exact structure, as each image corresponds to a unique rooted tree representation of region containment.

\paragraph{Topology-aware vision.}
Topology has been incorporated into vision models primarily through continuous relaxations. For example, topology-preserving losses enforce constraints in segmentation by matching Betti-number structure via persistent homology \cite{hu2019topoloss}. These approaches capture coarse invariants such as connectivity or holes at the pixel level. 

In contrast, CurveBench targets a discrete combinatorial object: the containment tree induced by disjoint Jordan curves. This representation encodes fine-grained nesting relationships and is closely related to classical diagrammatic representations such as Euler diagrams \cite{stapleton2014eulersurvey}. As such, our setting focuses on exact topology inference rather than topology regularization.

\paragraph{Reinforcement learning for structured reasoning.}
Our fine-tuning setup is motivated by recent work showing that reinforcement learning with verifiable rewards can improve structured reasoning without requiring human preference labels or annotated reasoning traces. Reinforcement Learning with Verifiable Rewards (RLVR) replaces learned reward models with deterministic reward functions computed from ground-truth verification, making it particularly suitable for tasks with objectively checkable outputs such as mathematics, code, and structured prediction~\cite{lambert2024tulu3}. This paradigm was further popularized by DeepSeekMath, which introduced Group Relative Policy Optimization (GRPO) for mathematical reasoning~\cite{deepseekmath2024grpo}, and by DeepSeek-R1, which showed that large-scale RL post-training with verifiable rewards can elicit stronger reasoning behavior in language models~\cite{deepseek2025r1}.

Recent work has also extended R1-style and RLVR-style training to vision-language models. VLM-R1 studies rule-based reinforcement learning for visual reasoning tasks and shows that verifiable visual tasks can benefit from RL-style post-training~\cite{shen2025vlmr1}. Similarly, LMM-R1 applies rule-based reinforcement learning to multimodal reasoning in small large multimodal models~\cite{peng2025lmmr1}, while R1-VL introduces a step-wise GRPO variant for multimodal reasoning~\cite{zhang2025r1vl}. Other recent work, such as Perception-R1 and MM-Eureka, further explores rule-based reinforcement learning for visual perception and multimodal reasoning tasks~\cite{yu2025perceptionr1,meng2025mmeureka}. CurveBench follows this direction but focuses on a different kind of visual reasoning: recovering a discrete topological structure from an image. Because the target output is a rooted containment tree, correctness can be evaluated exactly, enabling direct optimization of the task metric.

Our optimization objective builds on GRPO-style group-relative updates, but we use Dr.GRPO~\cite{liu2025understandingr1zero}, which identifies and corrects biases in the original GRPO objective. In particular, Dr.GRPO addresses issues such as biased advantage normalization and length-related effects that can distort optimization. This is relevant in our setting because outputs are structured and may vary in length depending on the predicted tree.

\paragraph{Parameter-efficient RL fine-tuning.}
To make RL post-training feasible for open-weight vision-language models, we use Low-Rank Adaptation (LoRA)~\cite{hu2022lora}. LoRA freezes the base model and trains a small set of low-rank adapter parameters, substantially reducing memory and compute requirements. This is especially appropriate for RL fine-tuning, where each rollout provides only a sparse, outcome-level learning signal rather than token-level supervision. The ``LoRA Without Regret'' study argues that RL updates often contain far less information per episode than supervised fine-tuning, and shows that sufficiently configured LoRA adapters can approach full fine-tuning performance in RL settings~\cite{schulman2025lora}. In CurveBench, each rollout is evaluated using two binary verifiable signals, tree correctness and node-count correctness, which further supports the use of a compact low-rank adaptation scheme.

\paragraph{Positioning.}
Overall, CurveBench occupies a unique point in the landscape: it combines vision-to-structure prediction, diagram-like controlled inputs, and exact verifiable evaluation. Unlike prior work that emphasizes semantic graphs, geometric reconstruction, or approximate topology, our benchmark isolates topological hierarchy extraction as a standalone capability, providing a controlled setting for evaluating and improving structure-aware visual reasoning.

\section{Dataset of CurveBench}

\begin{figure}[htbp]
     \centering
     \begin{subfigure}[b]{0.16\textwidth}
         \centering
         \includegraphics[width=\textwidth]{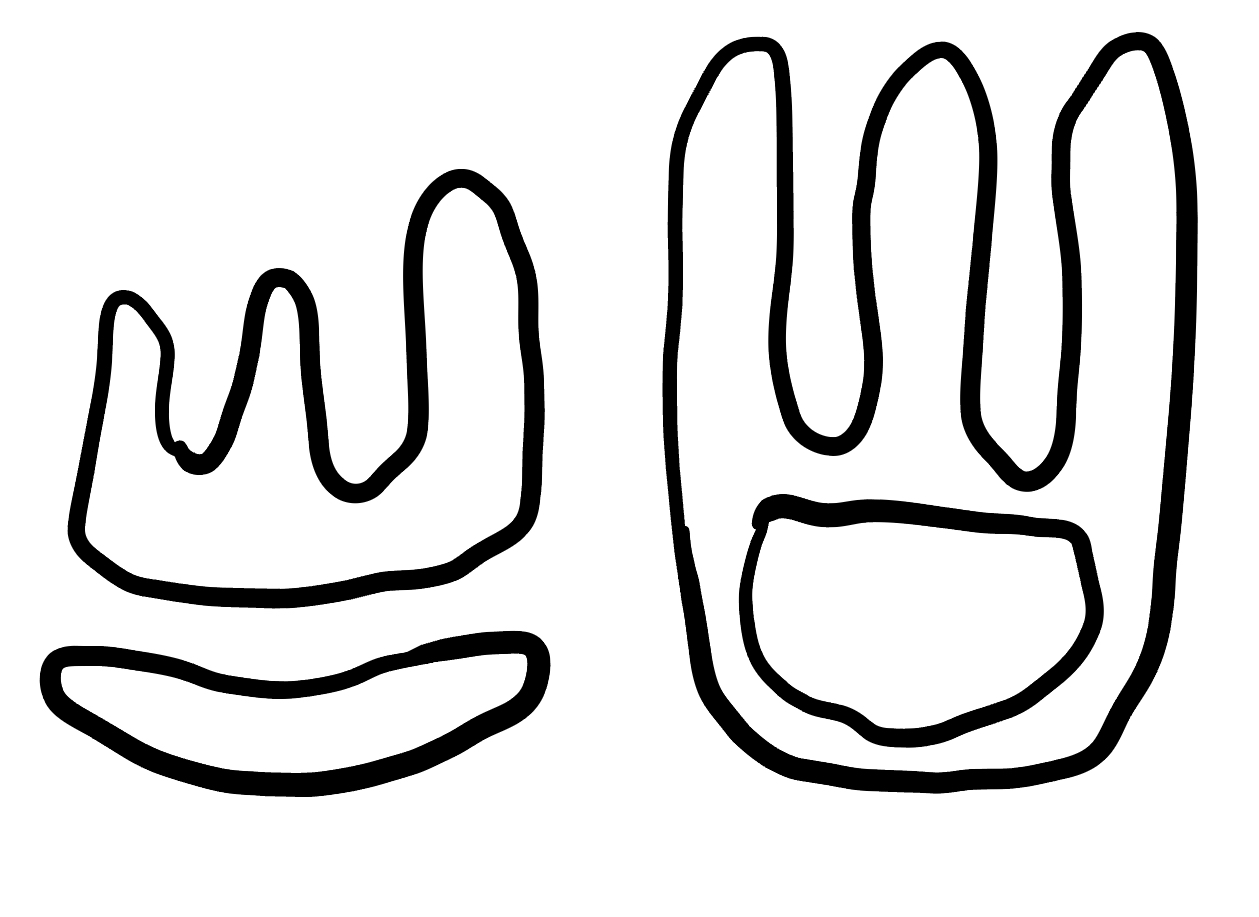}
         \caption{Easy}
         \label{fig:img2}
     \end{subfigure}
     \hfill
     \begin{subfigure}[b]{0.16\textwidth}
         \centering
         \includegraphics[width=\textwidth]{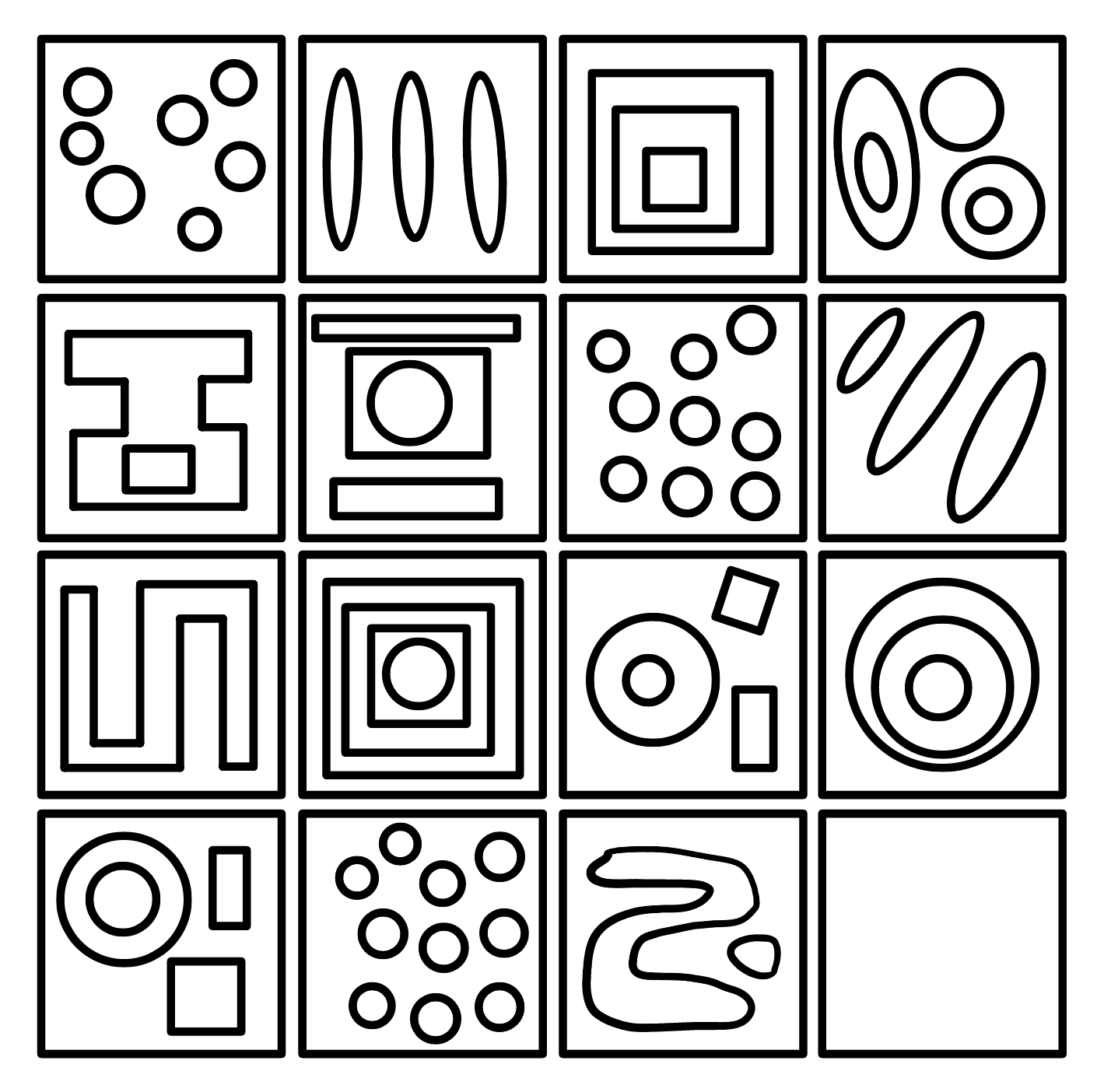}
         \caption{Counting}
         \label{fig:img3}
     \end{subfigure}
     \hfill
     \begin{subfigure}[b]{0.16\textwidth}
         \centering
         \includegraphics[width=\textwidth]{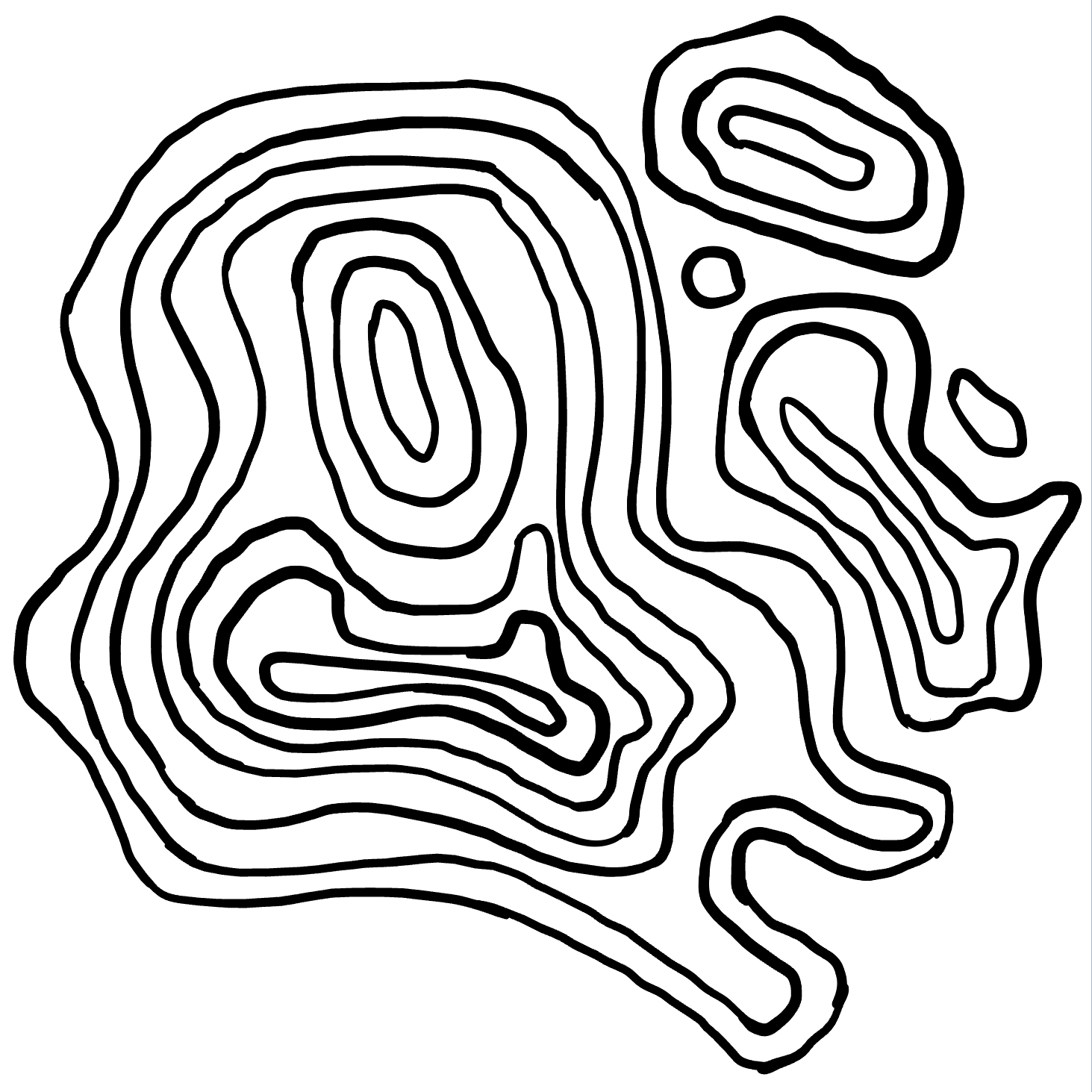}
         \caption{Topographical}
         \label{fig:img4}
     \end{subfigure}
     \hfill
     \begin{subfigure}[b]{0.16\textwidth}
         \centering
         \includegraphics[width=\textwidth]{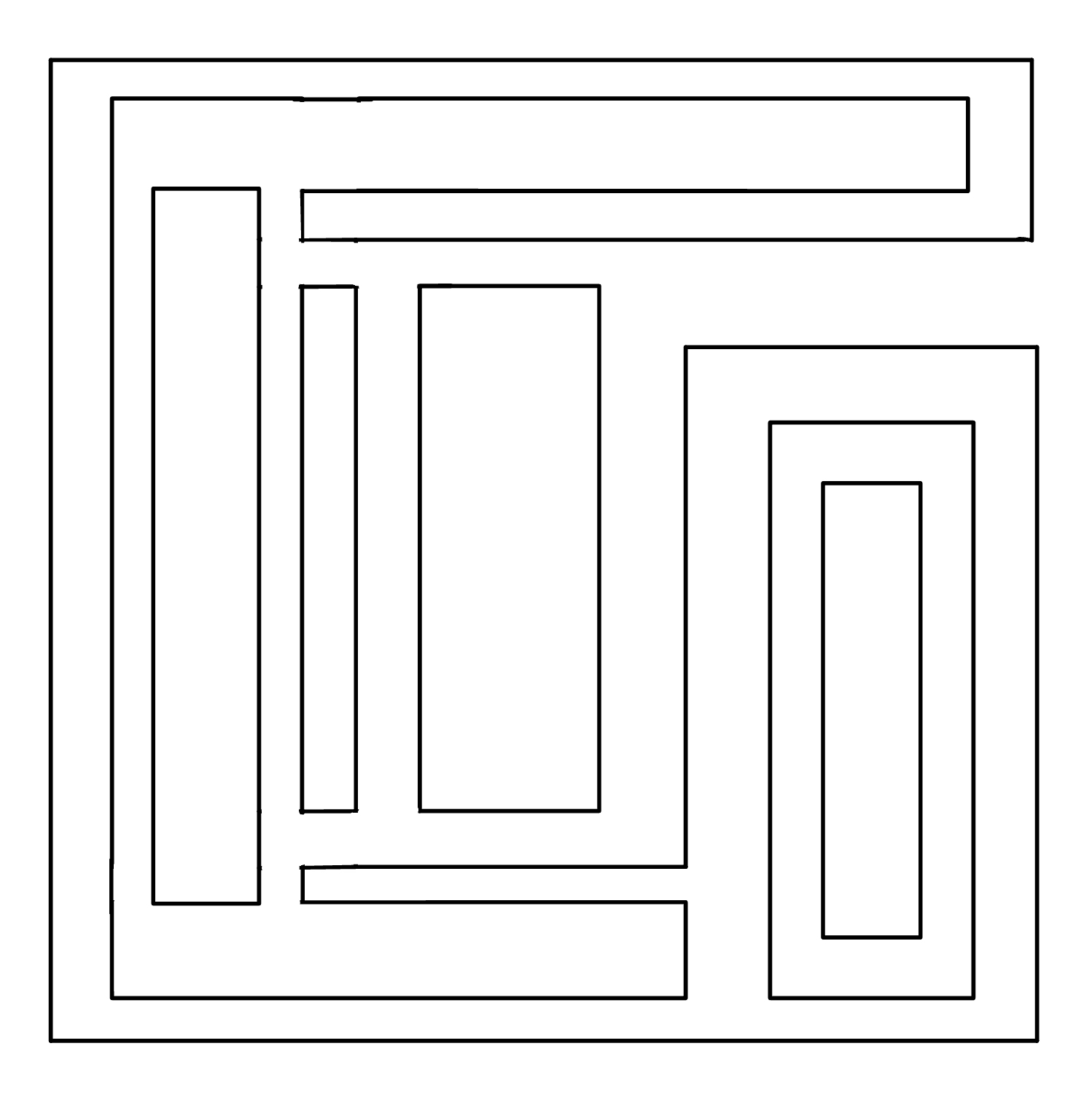}
         \caption{Maze}
         \label{fig:img5}
     \end{subfigure}
        \hfill
     \begin{subfigure}[b]{0.16\textwidth}
         \centering
         \includegraphics[width=\textwidth]{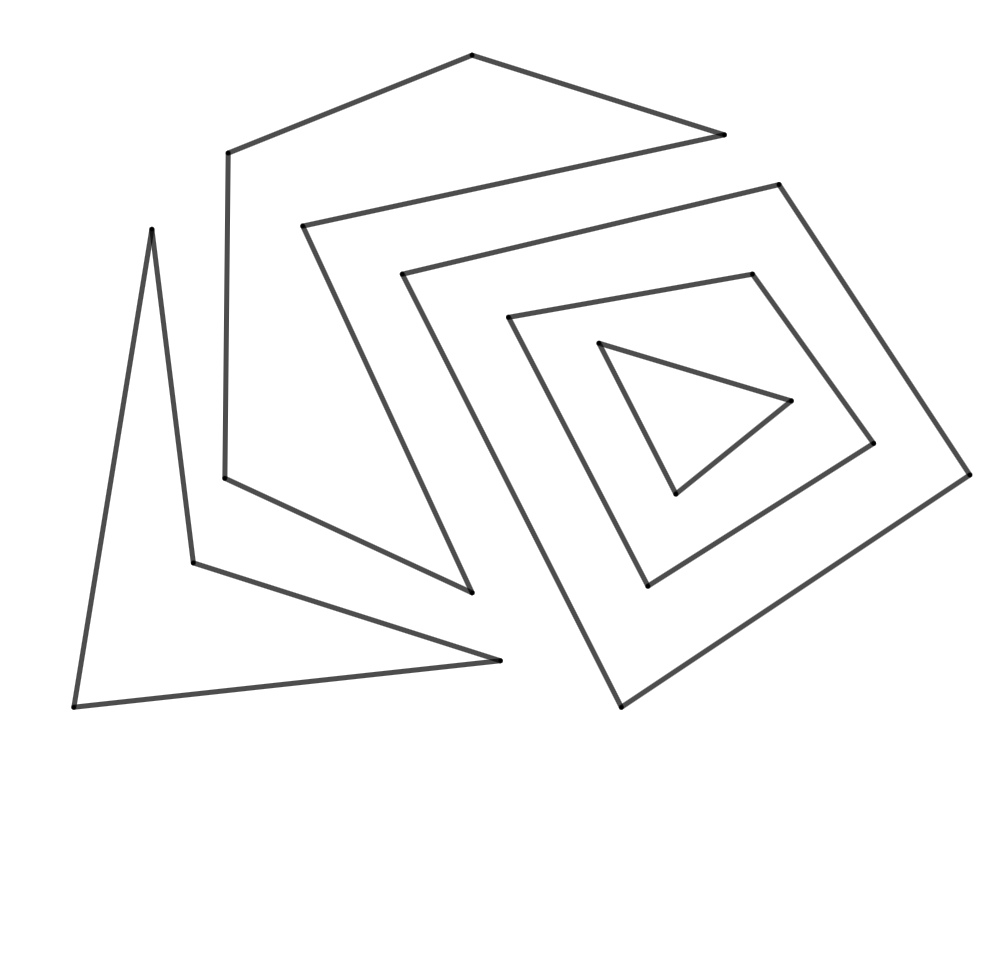}
         \caption{Polygon}
         \label{fig:img6}
     \end{subfigure}
        \caption{Representative examples from each category within the CurveBench dataset}
        \label{fig:five_images_row}
\end{figure}

CurveBench focuses specifically on exact recovery of rooted containment trees from images of pairwise disjoint Jordan curves by mapping visual containment to exact combinatorial structures. While existing datasets often evaluate semantic segmentation or geometric object detection, CurveBench isolates containment and separation as the core signals for visual reasoning. It requires models to infer a rooted tree whose nodes represent connected regions and whose edges denote the separating boundary curves. The dataset contains a total of 756 hand-drawn images, providing structural diversity without relying exclusively on a procedural rendering pipeline. Its 756-image scale and restriction to pairwise non-intersecting closed curves are deliberate: CurveBench is intended as a controlled diagnostic benchmark rather than a broad-coverage corpus. Figure~\ref{fig:five_images_row} shows representative examples.

    \textbf{Easy (300 images):} This subset establishes a fundamental baseline, containing relatively small curve configurations. The construction enumerates rooted tree structures and manually renders multiple examples of each selected structure. For each selected combinatorial tree, we manually authored at least two distinct visual representations. The Easy subset is further split into 210 training images, 45 validation images, and 45 held-out test images. The training and validation splits are used for RL fine-tuning, while the test split is reserved exclusively for final evaluation.

   \textbf{Polygon (199 images):} Following a systematic construction methodology identical to the Easy category, this subset restricts the geometries entirely to non-intersecting polygons. This tests a model's robustness to sharp angles and piecewise-linear boundaries compared to smooth, continuous Jordan curves.
   
    \textbf{Topographical (100 images):} Grounded in applied distributions, these images are directly inspired by real-world topographical maps. They mimic the natural behavior of elevation level sets, extending the evaluation from theoretical combinatorial benchmarks to practical visual understanding domains. The images in this subset are manually authored and original creations. While they are qualitatively inspired by the morphology of real-world elevation level sets, they do not contain data from external mapping services.
    
    \textbf{Maze (100 images):} Designed to stress-test long-range spatial reasoning, this category features highly convoluted, labyrinthine curves with deep nesting.  The spatial entanglement makes distinguishing the interior from the exterior of a boundary visually demanding, forcing models to track complex geometric boundaries over long distances.
    
 \textbf{Counting (57 images):} This densely populated subset evaluates a model's scalability and capacity limits. Focused primarily on the volume of nested entities, these images are packed with a high number of disjoint curves, challenging the framework to construct larger rooted trees without accumulating structural or logical errors.

The combined subset of Polygon, Topographical, Maze, and Counting images forms CurveBench-Hard, containing 456 images in total. Each image in CurveBench is paired with a formal combinatorial rooted tree representing the nesting structure of its planar regions. This annotation format enables deterministic evaluation of structural predictions, where models are assessed on their ability to exactly reconstruct the adjacency and containment relationships present in the visual input.
\paragraph{Ground-truth annotation.}
Ground-truth trees were initialized using an OpenCV contour-based extraction pipeline. The four authors divided the dataset into disjoint subsets. One annotator manually checked the regions and immediate parent--child relations in each image.

\section{Tree generation task}
\label{sec:treetask}
We formulate the nestedness extraction task as a structured prediction problem that maps an image of disjoint Jordan curves to its underlying topological hierarchy. Given an input image, the objective is to recover the containment relations between regions induced by the curves. More formally, the task is defined by the following inputs and outputs:

\textbf{Input.} An image containing a collection of pairwise disjoint Jordan curves in the plane. The curves may vary in shape, scale, and complexity, and may exhibit nested, adjacent, or maze-like configurations.

\textbf{Output.} A rooted tree representing the nestedness structure of the image. Each node corresponds to a region in the planar subdivision induced by the curves, and each edge represents an immediate containment relation, encoded by a shared boundary curve between two regions.

This formulation isolates topological structure as the primary prediction target and enables evaluation using tree-based structural metrics. We evaluate all models using a fixed instruction prompt that asks the model to output the rooted containment tree as a list of parent--child edges inside \texttt{<answer>} tags. The first line specifies the number of non-root nodes, and each subsequent line specifies an edge \texttt{u v}, where \texttt{v} is the parent of \texttt{u}. The full evaluation prompt is provided in Appendix~\ref{app:eval-prompt}.

Table~\ref{tab:mapping} shows a sample input, its corresponding tree, and the representation of the tree as the expected output.
\begin{table}[h]
\centering
\caption{Topological Mapping of Regions to Tree Structure}
\label{tab:mapping}
\begin{tabularx}{\textwidth}{@{}>{\centering\arraybackslash}X >{\centering\arraybackslash}X >{\centering\arraybackslash}m{0.09\textwidth}@{}}
\toprule
\textbf{Input} & \textbf{Corresponding tree} & \textbf{Output} \\
\midrule
\includegraphics[width=0.98\linewidth]{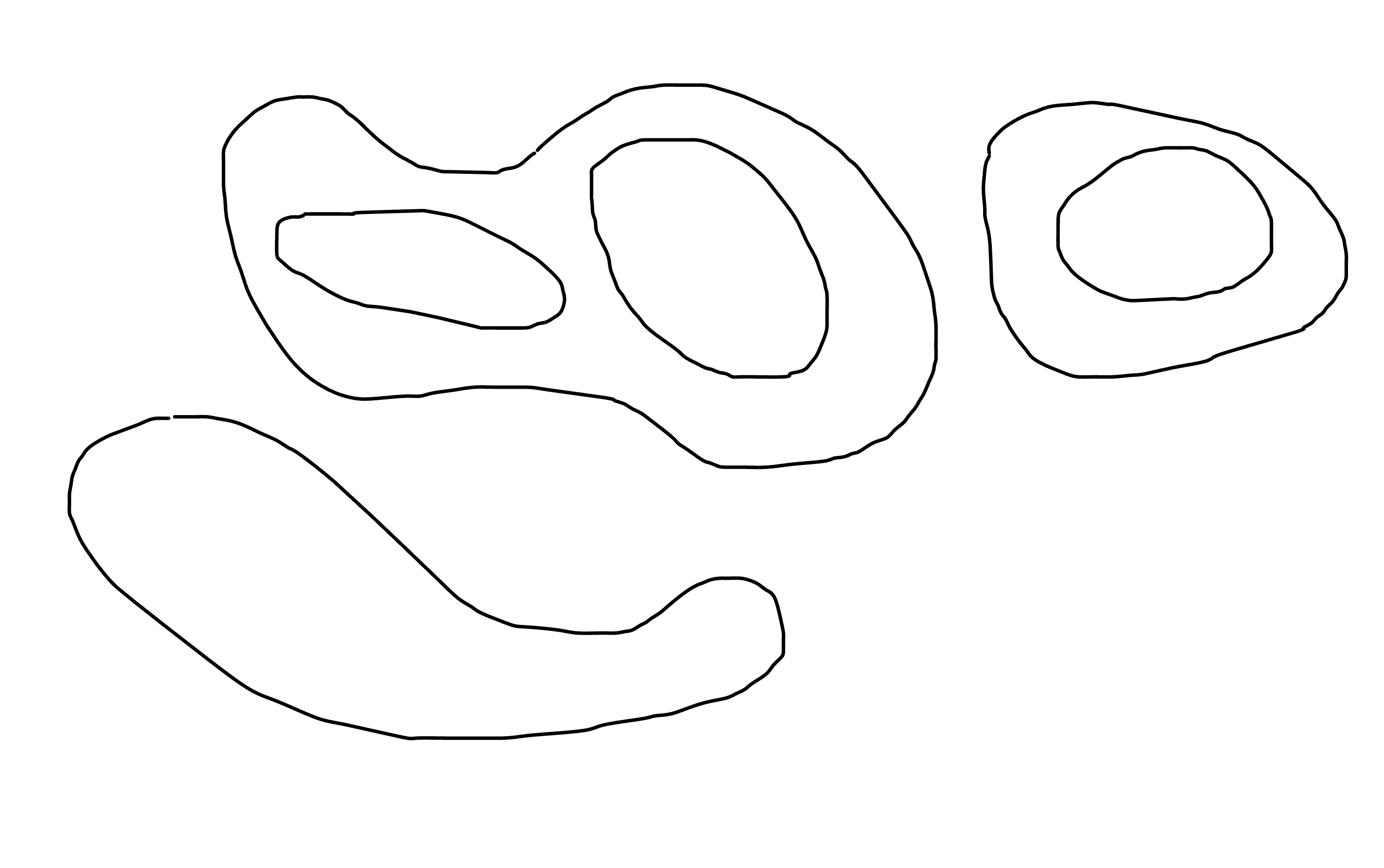} &
\includegraphics[width=0.98\linewidth]{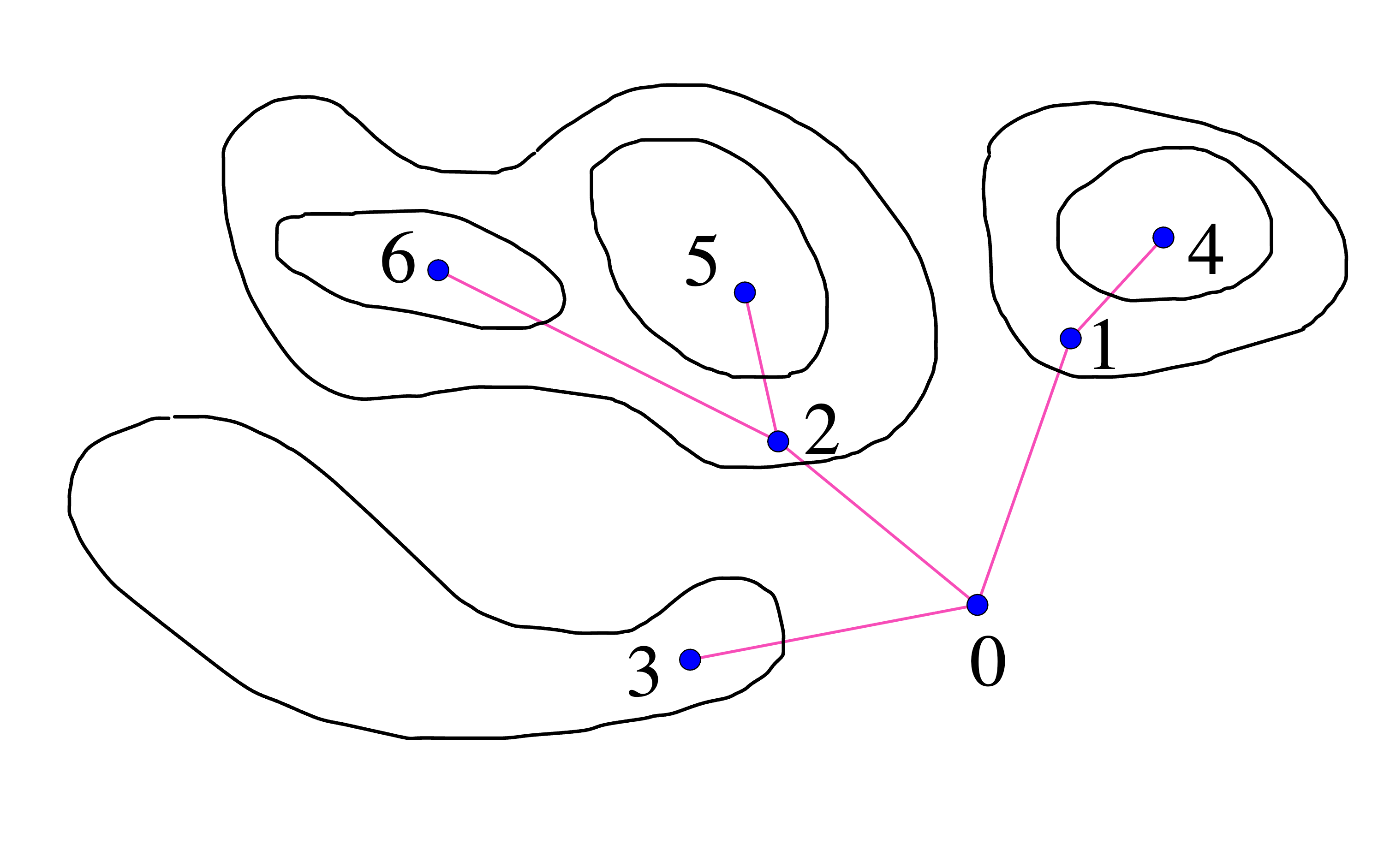} &
\includegraphics[width=0.65cm]{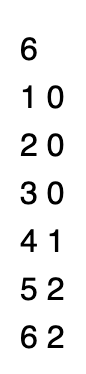} \\
\bottomrule
\end{tabularx}
\end{table}

\section{Experimental Setup}
\label{sec:experiments}

We improve structured topological prediction on CurveBench via reinforcement learning (RL) fine-tuning of open-weight VLMs. The fine-tuning experiments use the training and validation splits of CurveBench-Easy; the CurveBench-Easy test split is held out and is not used during training or model selection. Once training is complete, we evaluate all trained models and comparison models on the held-out CurveBench-Easy test split and on the full CurveBench-Hard benchmark. The former measures generalization within the easier distribution, while the latter measures transfer to more challenging curve configurations.

\paragraph{Base Models.}
We fine-tune two pretrained vision-language models:
\begin{itemize}
    \item \textbf{Qwen3-VL-8B Thinking}, from the Qwen3-VL family~\cite{bai2025qwen3vl}.
    \item \textbf{Gemma 3 12B Instruct}, from the Gemma family of open models~\cite{gemma32025}.
\end{itemize}

\paragraph{Reinforcement Learning Fine-Tuning.}
We fine-tune both models using Reinforcement Learning with Verifiable Rewards (RLVR)~\cite{lambert2024tulu3}. CurveBench is naturally suited to RLVR because the correctness of a generated containment tree can be evaluated deterministically against the ground truth. We use RLVR rather than SFT alone to optimize task-level correctness over sampled solutions, allowing the model to explore and reinforce successful solution trajectories instead of imitating a single reference response. The verifier and reward components are detailed below.

Policy optimization is performed using Dr.GRPO~\cite{liu2025understandingr1zero}, a variant of GRPO~\cite{deepseekmath2024grpo,deepseek2025r1}. For each input image, we sample multiple candidate outputs and compute their relative advantages from the corresponding verifier rewards. Dr.GRPO mitigates biases in the original GRPO objective, including length-related effects that are particularly relevant when structured outputs vary in textual length.

\paragraph{Evaluation rewards.}
The evaluation uses two binary reward components computed from the prediction.

\begin{itemize}
    \item \textbf{Node Count Accuracy (30\% weight):}
    \(R_{\text{count}} = 1\) if the predicted number of non-root nodes exactly matches the ground-truth count, and \(0\) otherwise.

    \item \textbf{Tree Structure Accuracy (70\% weight):}
    \(R_{\text{tree}} = 1\) if the predicted rooted tree exactly matches the ground-truth containment structure, and \(0\) otherwise.
\end{itemize}

The combined reward is
\[
R_{\mathrm{comb}} = 0.3 \cdot R_{\text{count}} + 0.7 \cdot R_{\text{tree}}.
\]

Since both reward components are binary, the combined reward can take only four possible values:
\(
R_{\mathrm{comb}} \in \{0, 0.3, 0.7, 1.0\}.
\)
Thus, each rollout provides a sparse outcome-level signal rather than dense token-level supervision. This makes CurveBench well-suited to RLVR: correctness can be checked exactly, but the learning signal is minimal.

We compare two Qwen3-VL-8B Thinking checkpoints, a \textbf{combined-reward variant} trained with node-count and tree-structure rewards and a \textbf{tree-only variant} trained with the tree-structure reward alone. We evaluate both checkpoints on the same held-out data using rooted-tree accuracy, node-count accuracy, and the combined evaluation reward. Rooted-tree accuracy is the primary metric.

\paragraph{Root-preserving tree matching.}
The exterior region is the distinguished root, labeled $0$. Exact correctness requires an isomorphism $\phi$ between the predicted and reference trees with \textbf{$\phi(0)=0$}; other labels and sibling order may vary. The evaluation verifier validates directed parent--child trees and compares recursively sorted child-subtree signatures starting at $0$. Missing or invalid trees receive zero tree reward. The model's text format lists \texttt{child parent}; the parser converts it to internal \texttt{(parent, child)} edges.

For example, $\{(0,1),(1,2)\}$ describes two nested curves, whereas $\{(0,1),(0,2)\}$ describes two separate curves. Their unrooted graphs are isomorphic, but their rooted trees differ because the exterior has one child versus two. This prediction receives a tree reward of zero and a count reward of one.

\paragraph{Checkpoint evaluation.}
The released checkpoints were trained with a tree reward based on unrooted graph isomorphism. We score their archived completions with the root-preserving verifier. Appendix~\ref{app:code-artifacts} lists the software and checkpoint identifiers.

\paragraph{Parameter-Efficient Fine-Tuning.}
We employ Low-Rank Adaptation (LoRA)~\cite{hu2022lora} for parameter-efficient RL fine-tuning. Only LoRA adapter parameters are updated, while the base model weights remain frozen. We use the \texttt{all-linear} target-module configuration in TRL, which applies adapters to linear layers throughout the model rather than restricting adaptation to a small subset of modules. This provides broad adaptation capacity while substantially reducing memory usage and training cost compared to full fine-tuning.

LoRA is particularly suitable for our RLVR setting because the verifier provides sparse outcome-level feedback rather than dense token-level supervision. Each rollout receives binary feedback for tree correctness and node-count correctness, combined into one of four possible reward values. Prior empirical work on LoRA-based RL fine-tuning suggests that appropriately configured adapters can approach full fine-tuning performance in such low-information RL settings~\cite{schulman2025lora}. We therefore use a compact LoRA configuration with rank \(r=4\) and scaling factor \(\alpha=8\).

\paragraph{Training Configuration.}
Models are trained for 250 optimization steps with a batch size of 128 and 8 sampled generations per input. A constant learning rate of $8 \times 10^{-5}$ is used throughout training. We fine-tuned the models on eight NVIDIA RTX PRO 6000 GPUs.

\paragraph{Evaluation environment.}
We evaluated the fine-tuned checkpoints on two NVIDIA RTX PRO 6000 GPUs and evaluated the remaining models through the Prime Intellect Environments Hub~\cite{primeintellect_envhub}. The evaluation protocol for each dataset specifies the split, input format, prompt, answer parser, and reward function. Appendix~\ref{app:environments} lists the released environments for CurveBench-Easy and CurveBench-Hard.

\section{Results}
\label{sec:results}

Tables~\ref{tab:curvebench-easy-results-brief} and~\ref{tab:curvebench-hard-results-brief} report rooted-tree accuracy, node-count accuracy, and combined reward for all available runs. CurveBench-Easy contains 45 held-out images with four rollouts each, yielding 180 evaluated responses per model. CurveBench-Hard contains 456 images with one rollout each. We average the per-response scores; these are not best-of-four or pass@4 results. All values lie in $[0,1]$, and combined reward is $0.7$ times tree accuracy plus $0.3$ times count accuracy, computed before rounding. Bold entries are column maxima. We report descriptive point estimates; repeated rollouts on one image are not independent images.

\begin{table}[htbp]
\centering
\normalsize
\setlength{\tabcolsep}{6pt}
\renewcommand{\arraystretch}{1.10}
\begin{tabularx}{\textwidth}{@{}Xrrr@{}}
\toprule
Model & Rooted tree & Node count & Avg. reward \\
\midrule
Gemini 3.1 Pro Preview & \textbf{0.711} & \textbf{0.778} & \textbf{0.731} \\ 
Gemini 3 Pro Preview & 0.639 & 0.739 & 0.669 \\ 
GPT-5.2 & 0.389 & 0.433 & 0.402 \\ 
Qwen3-VL-235B-A22B Thinking & 0.333 & 0.522 & 0.390 \\ 
\rowcolor{blue!10} Qwen3-VL-8B (combined reward) & 0.317 & 0.544 & 0.385 \\ 
GPT-5.4 & 0.306 & 0.422 & 0.341 \\ 
Claude Opus 4.5 & 0.294 & 0.433 & 0.336 \\ 
\rowcolor{blue!10} Qwen3-VL-8B (tree-only) & 0.244 & 0.494 & 0.319 \\ 
\rowcolor{orange!10} Qwen3-VL-8B Thinking (16K) & 0.217 & 0.400 & 0.272 \\ 
GPT-5 Mini & 0.172 & 0.200 & 0.181 \\ 
GPT-5.4 Mini & 0.100 & 0.383 & 0.185 \\ 
\rowcolor{blue!10} Gemma-3-12B (combined reward) & 0.078 & 0.489 & 0.201 \\ 
Gemma-3-27B-IT & 0.039 & 0.278 & 0.111 \\ 
Qwen3-VL-8B Thinking (8K) & 0.028 & 0.061 & 0.038 \\ 
\rowcolor{orange!10} Gemma-3-12B-IT & 0.022 & 0.233 & 0.086 \\ 
Qwen3-VL-8B Instruct & 0.017 & 0.322 & 0.108 \\ 
\bottomrule
\end{tabularx}
\caption{CurveBench-Easy results on 45 held-out images, with four rollouts per image (180 responses per model). Values are response-level accuracies/rewards, not pass@4. All runs are sorted by rooted-tree accuracy. Blue rows are released fine-tuned checkpoints; orange rows are their principal comparison bases. Thinking (8K) and Thinking (16K) denote distinct completion-budget runs.}
\label{tab:curvebench-easy-results-brief}
\label{tab:curvebench-easy-results}
\end{table}

\begin{table}[htbp]
\centering
\normalsize
\setlength{\tabcolsep}{6pt}
\renewcommand{\arraystretch}{1.10}
\begin{tabularx}{\textwidth}{@{}Xrrr@{}}
\toprule
Model & Rooted tree & Node count & Avg. reward \\
\midrule
Gemini 3.1 Pro Preview & \textbf{0.287} & \textbf{0.445} & \textbf{0.335} \\ 
Gemini 3 Pro Preview & 0.254 & 0.404 & 0.299 \\ 
Gemini 3 Flash Preview & 0.154 & 0.287 & 0.194 \\ 
GPT-5.2 & 0.123 & 0.191 & 0.143 \\ 
GPT-5.4 & 0.105 & 0.221 & 0.140 \\ 
\rowcolor{blue!10} Qwen3-VL-8B (combined reward) & 0.090 & 0.232 & 0.133 \\ 
Qwen3-VL-235B-A22B Thinking & 0.086 & 0.243 & 0.133 \\ 
\rowcolor{orange!10} Qwen3-VL-8B Thinking (16K) & 0.079 & 0.208 & 0.118 \\ 
Claude Opus 4.5 & 0.070 & 0.171 & 0.100 \\ 
\rowcolor{blue!10} Qwen3-VL-8B (tree-only) & 0.057 & 0.226 & 0.108 \\ 
Qwen3-VL-8B Thinking (8K) & 0.046 & 0.121 & 0.068 \\ 
GPT-5 Mini & 0.033 & 0.090 & 0.050 \\ 
GPT-5.4 Mini & 0.029 & 0.132 & 0.059 \\ 
Qwen3-VL-8B Instruct & 0.024 & 0.162 & 0.066 \\ 
\rowcolor{blue!10} Gemma-3-12B (combined reward) & 0.015 & 0.237 & 0.082 \\ 
\rowcolor{orange!10} Gemma-3-12B-IT & 0.009 & 0.081 & 0.030 \\ 
\bottomrule
\end{tabularx}
\caption{CurveBench-Hard results on all 456 images, with one rollout per image. Rows are sorted by rooted-tree accuracy. Colors and completion-budget labels follow Table~\ref{tab:curvebench-easy-results-brief}.}
\label{tab:curvebench-hard-results-brief}
\label{tab:curvebench-hard-results}
\end{table}

\paragraph{Easy performance.}
Gemini 3.1 Pro leads all three metrics, with 0.711 rooted-tree accuracy. The Qwen3-VL-8B combined-reward checkpoint reaches 0.317 tree accuracy and 0.385 combined reward, compared with 0.217 and 0.272 for its 16K Thinking base. The tree-accuracy difference is 10.0 percentage points (57 versus 39 correct responses out of 180). The trained checkpoint is numerically above GPT-5.4 (0.306) and Claude Opus 4.5 (0.294), with no claim of statistical significance. The tree-only checkpoint reaches 0.244 tree accuracy.

\paragraph{Hard transfer and completion budget.}
On CurveBench-Hard, Gemini 3.1 Pro reaches 0.287 tree accuracy. The combined-reward Qwen checkpoint reaches 0.090, compared with 0.079 for its 16K base: a difference of five correct predictions out of 456 (1.1 percentage points). The tree-only checkpoint scores 0.057, below the 16K base. Hard transfer therefore remains limited and is not uniform across training variants. Increasing the Thinking completion budget from 8K to 16K raises tree accuracy from 0.028 to 0.217 on Easy and from 0.046 to 0.079 on Hard; combined reward changes from 0.038 to 0.272 and from 0.068 to 0.118, respectively. Appendix~\ref{app:thinking-budget} gives the complete comparison.

\FloatBarrier

\section{Limitations}
\label{sec:limitations}

First, CurveBench is modest in size, consisting of 756 images. This scale is small relative to large general-purpose vision corpora, but it reflects a deliberate trade-off: the benchmark prioritizes high-quality structural annotations, human verification, and exact tree-based evaluation over dataset scale. CurveBench is therefore intended primarily as a diagnostic benchmark rather than a large-scale pretraining corpus.

Second, CurveBench focuses on a specific class of topological structures: nested, pairwise non-intersecting Jordan curves. This controlled setting enables deterministic evaluation of rooted containment trees, but it does not cover all forms of visual topology or spatial reasoning. The benchmark does not include intersecting curves, open contours, noisy real-world segmentations, three-dimensional topology, temporal structure, or natural images with ambiguous boundaries. Extending CurveBench to these settings would likely require different annotation schemes and evaluation metrics.

Third, training is currently limited to CurveBench-Easy because Hard often yields near-zero rewards, so our experiments primarily test Easy and only indirect Hard transfer. Parent-edge F1 and normalized isomorphic-prefix scores could provide partial rewards for Hard predictions that recover some of the structure. Future work will also remove intermediate-depth curves---whitening deeper inner curves---to form a restricted-complexity Hard curriculum before retraining.

Fourth, CurveBench is intentionally synthetic and controlled. This design enables exact ground-truth construction, deterministic evaluation, and clean isolation of region-containment reasoning, but it also limits ecological validity. The images are hand-drawn diagrams with controlled visual structure. As a result, strong performance on CurveBench should not be interpreted as sufficient evidence that a model can robustly handle natural maps, scientific figures, medical images, or arbitrary contour-like structures in the wild. Conversely, poor performance on CurveBench should be interpreted as evidence of difficulty with exact topological abstraction under controlled conditions, not as a complete measure of general visual intelligence. Future dataset extensions should introduce additional visual styles, rendering artifacts, ambiguous boundaries, intersections, open curves, and real-world contour sources while preserving verifiable structural annotations.

Finally, exact tree match gives the same score to every structurally incorrect prediction. Partial-credit metrics could distinguish degrees of structural agreement. Each image was checked by a single annotator, which precludes estimating inter-annotator agreement.

\section{Conclusion}
We introduced CurveBench, a benchmark for topology-aware visual reasoning in which a model must recover the exact rooted containment tree induced by an image of pairwise disjoint Jordan curves. Despite the visual simplicity of these inputs, our results show that exact hierarchical reconstruction remains challenging for current vision-language models, especially on structurally complex instances. The combined-reward Qwen3-VL-8B checkpoint improves over its base model on CurveBench-Easy and shows limited transfer to CurveBench-Hard.

More broadly, CurveBench highlights a capability that is largely orthogonal to object recognition and OCR: the ability to infer the global combinatorial organization of planar space. A natural next direction is to move beyond containment trees to full planar maps and their dual graphs. In our setting, the target rooted tree can already be viewed as the dual graph of the planar subdivision induced by the disjoint curves, rooted at the exterior face; extending this viewpoint to general planar subdivisions would require models to recover richer adjacency structure, including cycles and non-nested interactions between regions. Such a generalization would connect topology-aware vision to applications in cartography and GIS, map vectorization, scientific imaging, and structured scene understanding. We hope CurveBench serves not only as a benchmark for current systems, but also as a foundation for future models that can recover exact combinatorial structure from visual input.

\clearpage
\section*{Acknowledgements and Disclosures}
We thank Sara Javanmardi for initially introducing the team members. This research was partially supported by an unrestricted gift from Google. This funding did not grant Google any advance preview or editorial influence over the results. The work of Naser Talebizadeh Sardari was partially supported by NSF grant DMS-2401242.

\bibliographystyle{plainnat}
\bibliography{main}

\newpage

\appendix

\section{Prompts}
\label{app:prompts}

\subsection{Evaluation prompt}
\label{app:eval-prompt}

All models were evaluated using the same fixed prompt. The prompt asks the model to recover the rooted containment tree induced by the image and to return the answer in a parseable format. Its initial edge-order wording says parent--child, while the subsequent line-format instruction, example, and parser specify child--parent. The prompt below is the one used to generate the evaluated responses.

\begin{tcolorbox}[colback=gray!5, colframe=black!60, boxrule=0.5pt, arc=3pt]
\textbf{Prompt}: Analyze this image and extract the hierarchical tree structure representing the nested regions.

The image contains nested shapes/regions. Your task is to identify the parent-child relationships between these regions.

Return the tree structure as a list of edges, where each edge is represented as \texttt{(parent, child)}.

\begin{itemize}
    \item The root node is always \texttt{0}.
    \item Each region is assigned a unique node number.
    \item Edges represent parent-child relationships, where a parent region contains a child region.
\end{itemize}

Format your response inside \texttt{<answer>...</answer>} tags.

The first line should be the number of nodes excluding the root. Each subsequent line should be \texttt{u v}, meaning an edge from \texttt{v} to \texttt{u}, where \texttt{v} is the parent and \texttt{u} is the child.

Example:
\begin{verbatim}
<answer>
3
1 0
2 0
3 1
</answer>
\end{verbatim}

Make sure to include all edges that represent the hierarchical structure.
\end{tcolorbox}

\subsection{Prompt variants}
\label{app:prompt-variants}

Unless otherwise stated, the results reported in the main paper use the evaluation prompt in Appendix~\ref{app:eval-prompt}. We also considered variants that provide additional hints, such as the approximate number of nodes, but these are not used for the main benchmark results. This distinction is important because providing a node-count hint changes the task's difficulty and can affect both node-count accuracy and tree-structure accuracy.

\section{Datasets, Metadata, Environments, Code, and Artifacts}
\label{app:resources}

\subsection{Released Datasets}
\label{app:datasets}

\begin{wrapfigure}{r}{0.38\textwidth} 
    \centering
    \includegraphics[width=0.36\textwidth]{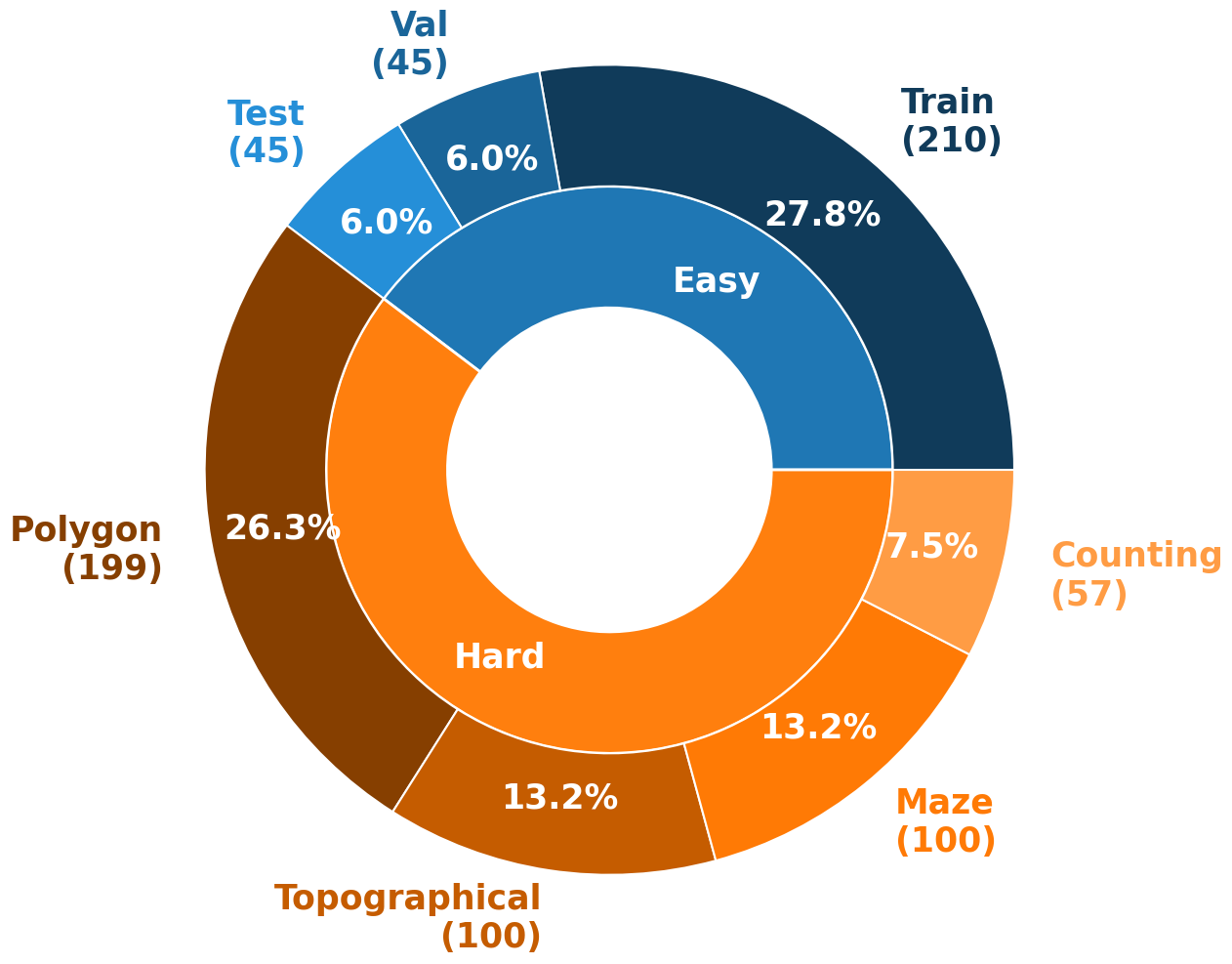}
    
    \captionsetup{justification=centering}
    
    \caption{CurveBench Dataset: \\ Hierarchical Distribution}
    \label{fig:wrapped_image}
\end{wrapfigure}
CurveBench is released as a collection of benchmark datasets for evaluating visual topological reasoning. The released resources include CurveBench-Easy and the main CurveBench benchmark used for the harder evaluation setting.

\paragraph{Public release.}
The public release includes the hosted datasets, Croissant metadata, evaluation environments, and code.

\begin{table}[h]
\centering
\small
\begin{tabularx}{\textwidth}{@{} l X @{}}
\toprule
\textbf{Resource} & \textbf{Description} \\
\midrule
CurveBench collection &
Collection containing the released CurveBench datasets and benchmark variants. \\
CurveBench-Easy &
Easy benchmark variant containing smaller curve configurations with simpler rooted containment trees. \\
CurveBench &
Main benchmark variant containing the harder evaluation categories used in our experiments. \\
Code repository &
Repository containing dataset construction utilities, ground-truth generation code, evaluation scripts, training code, experiment logs, and benchmark resources. \\ \\
\bottomrule
\end{tabularx}
\vspace{0.5em}
\caption{Dataset and code resources for CurveBench. Public resources support inspection of hosted datasets, Croissant metadata, evaluation environments, and code.}
\label{tab:dataset-resources}
\end{table}

The datasets, code, training artifacts, experiment logs, and ground-truth generation utilities are available at the following locations:
\begin{itemize}
    \item CurveBench collection: \url{https://huggingface.co/collections/AmirMohseni/curvebench}
    \item CurveBench-Easy: \url{https://huggingface.co/datasets/AmirMohseni/CurveBench-Easy}
    \item CurveBench: \url{https://huggingface.co/datasets/AmirMohseni/CurveBench}
    \item Code: \url{https://github.com/Amir-Mohseni/CurveBench}
\end{itemize}

The dataset is released under the \textbf{CC BY 4.0} license. The accompanying benchmark and evaluation code is released under the \textbf{MIT License}. These licenses are also specified in the dataset card, repository documentation, and Croissant metadata files.

\subsection{Croissant Metadata}
\label{app:croissant}

We include Croissant JSON-LD files with core and Responsible AI metadata for both datasets in the OpenReview submission.
\begin{itemize}
    \item CurveBench-Easy: \texttt{curvebench-easy-croissant.json}
    \item CurveBench: \texttt{curvebench-croissant.json}
\end{itemize}

\subsection{Evaluation Environments}
\label{app:environments}

We provide standardized evaluation environments that fix the dataset split, input formatting,
evaluation prompt, answer parser, and reward function. This ensures that different models are
evaluated under the same conditions and makes the reported benchmark results easier to reproduce.

\begin{table}[htbp]
\centering
\small
\begin{tabularx}{\textwidth}{@{} l X @{}}
\toprule
\textbf{Environment} & \textbf{Description} \\
\midrule
CurveBench-Easy &
Evaluation environment for the CurveBench-Easy test split. \\
CurveBench-Hard &
Evaluation environment for the full CurveBench-Hard benchmark set. \\
\bottomrule
\end{tabularx}
\caption{Evaluation environments used for CurveBench. Each environment specifies the dataset split, input format, evaluation prompt, answer parser, and reward function.}
\label{tab:environment-resources}

\end{table}

The aggregate results appear in Tables~\ref{tab:curvebench-easy-results} and~\ref{tab:curvebench-hard-results}.

\subsection{Category-Level and Partial-Credit Evaluation}
\label{app:category-analysis}
\label{app:structural-metrics}
We report aggregate scores for exact tree match, node count, and their combined reward. Category-level accuracies, joint reward distributions, and partial-credit metrics could show where models succeed and how they fail.

For future diagnostics, parent-edge F1 can be defined as the matched parent--child edge count divided by the pair's average edge count under a specified root-preserving matching. Normalized isomorphic-prefix depth measures the deepest depth through which rooted trees agree, divided by their maximum height, with explicit handling of root-only and invalid trees. Both require a documented parser, reference annotations, and matching implementation.

\subsection{Thinking-Model Token-Budget Analysis}
\label{app:thinking-budget}
We compare 8,192- and 16,384-token completion budgets for Qwen3-VL-8B Thinking with the same prompt, decoding settings, and rollout protocol. We score the archived completions with exact rooted-tree matching. The larger budget is associated with higher accuracy on both datasets in these runs.

\begin{table}[htbp]
\centering
\small
\setlength{\tabcolsep}{10pt}
\begin{tabular}{@{}lcc@{}}
\toprule
Dataset / metric & Thinking budget 8K & Thinking budget 16K \\
\midrule
Easy / Tree accuracy & 0.028 & 0.217 \\
Easy / Node accuracy & 0.061 & 0.400 \\
Easy / Combined reward & 0.038 & 0.272 \\
Hard / Tree accuracy & 0.046 & 0.079 \\
Hard / Node accuracy & 0.121 & 0.208 \\
Hard / Combined reward & 0.068 & 0.118 \\
\bottomrule
\end{tabular}

\caption{Comparison of 8K and 16K maximum completion lengths. Each run has 180 responses on Easy and 456 on Hard.}
\label{tab:thinking-budget}
\end{table}

\subsection{Training Diagnostics}
\label{app:historical-training}
\begin{figure}[htbp]
    \centering

    \begin{minipage}{0.48\textwidth}
        \centering
        \includegraphics[width=\linewidth]{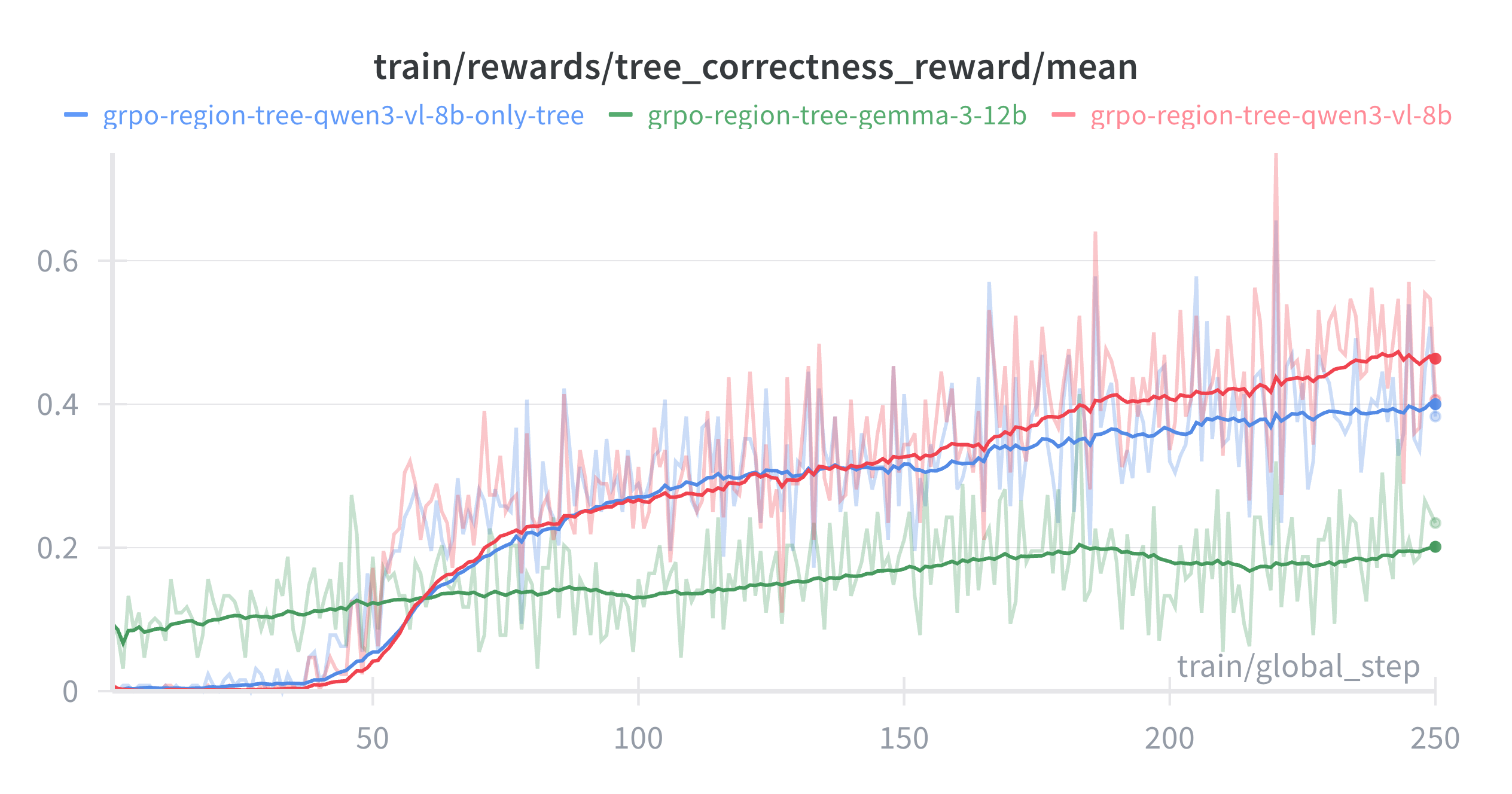}
    \end{minipage}
    \hfill
    \begin{minipage}{0.48\textwidth}
        \centering
        \includegraphics[width=\linewidth]{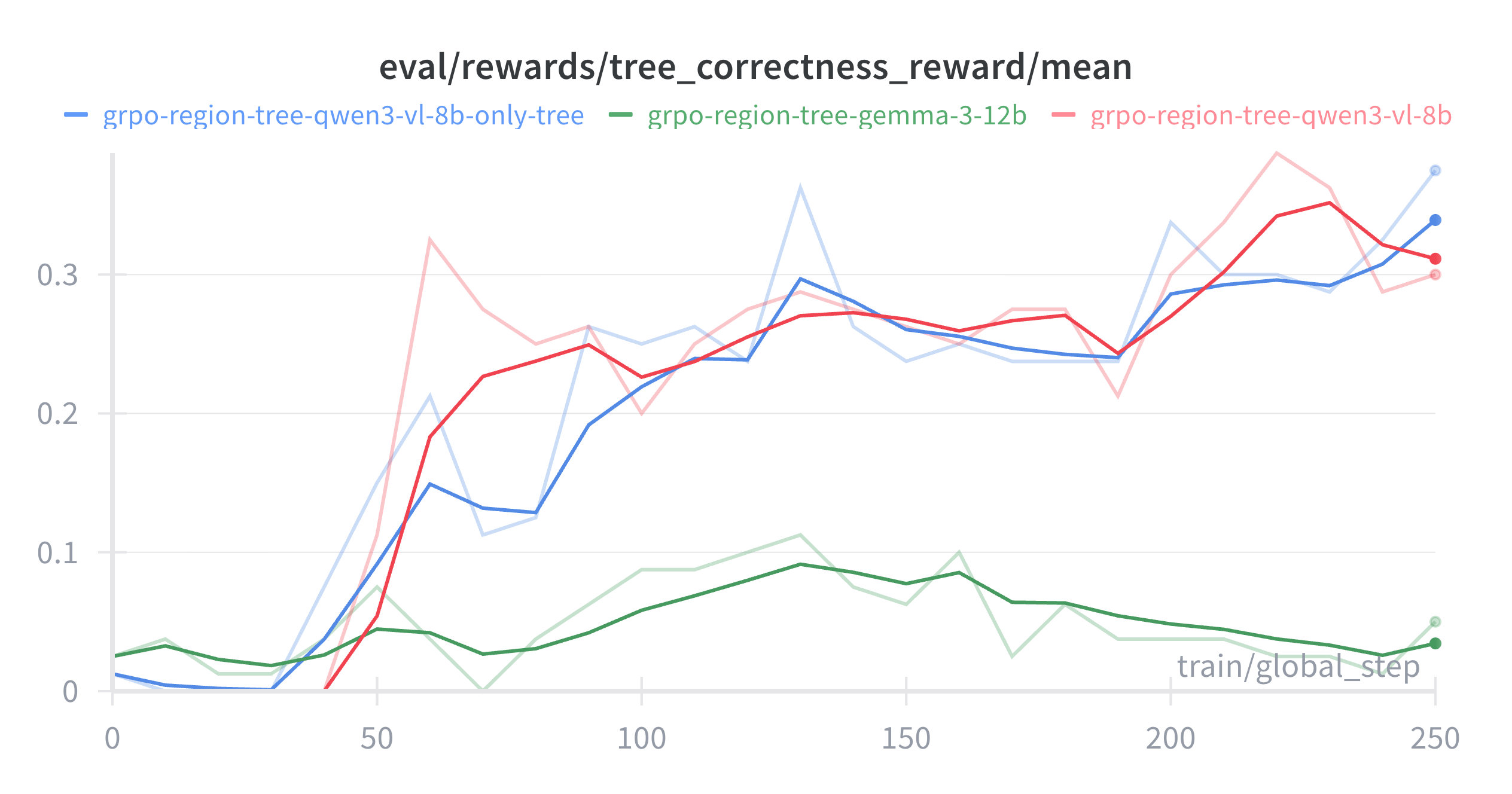}
    \end{minipage}

    \vspace{0.5em}

    \caption{Logged reward trajectories for training (left) and validation (right). Lighter curves show unsmoothed averages. Darker curves show a time-weighted exponential moving average with a factor of 0.98.}
    \label{fig:tree-reward-curves}
\end{figure}

Figure~\ref{fig:tree-reward-curves} shows the logged training and validation rewards, which use unrooted matching.

The evaluation environments are available through the public project resources:
\begin{itemize}
    \item CurveBench-Easy: \url{https://app.primeintellect.ai/dashboard/environments/amirmohseni/curvebench-env}
    \item CurveBench-Hard: \url{https://app.primeintellect.ai/dashboard/environments/amirmohseni/curvebench-hard-env}
\end{itemize}
\FloatBarrier

\subsection{Code, Training Artifacts, Logs, and Ground-Truth Generation}
\label{app:code-artifacts}

The public CurveBench resources include dataset construction utilities, OpenCV-based ground-truth extraction scripts, evaluation parsers, reward computation code, benchmark environment resources, reinforcement-learning training code, and training/evaluation logs.

Ground-truth trees were initialized using an automated OpenCV contour-based extraction pipeline that traces boundaries and assembles a containment tree rooted at the exterior. The four authors manually verified disjoint subsets, with one annotator per image.

\paragraph{Run identifiers.}
The fine-tuned rows retain run identifiers \texttt{qwen3-vl-8b-region-tree}, \texttt{qwen3-vl-8b-only-tree}, and \texttt{gemma-3-12b-region-tree}. Their public checkpoints are \nolinkurl{AmirMohseni/curvebench-qwen3-vl-8b}, \nolinkurl{AmirMohseni/curvebench-qwen3-vl-8b-only-tree}, and \nolinkurl{AmirMohseni/curvebench-gemma-3-12b}, respectively. The saved 8K and 16K Qwen Thinking runs use case-distinct identifiers \nolinkurl{Qwen/Qwen3-VL-8B-Thinking} and \nolinkurl{qwen/qwen3-vl-8b-thinking}; they are treated as separate runs of the same base model.

\paragraph{Software versions.}
The root-preserving comparator is present in the main repository at commit \texttt{53222e66cd5e} and in the training repository's \texttt{curvebench} branch at commit \texttt{a13f9bb8cd5c}; the main repository pins that training commit as a submodule. The published evaluation packages inspected were \texttt{curvebench-env==0.1.4} (Easy) and \texttt{curvebench-hard-env==0.1.2} (Hard). Each compares rooted signatures starting at node $0$ and rejects the nested-versus-separate counterexample. These identifiers locate the public implementations.

Reproducing the reported aggregates exactly requires the dataset revisions, evaluator builds, and archived completions used for scoring. The aggregate evaluation records do not identify all of these inputs, and the evaluation loaders do not pin dataset revisions.

Public code and training sources:
\begin{itemize}
    \item \url{https://github.com/Amir-Mohseni/CurveBench}
    \item \url{https://github.com/AmirTuring/trl/tree/curvebench}
\end{itemize}

\section{Ethical Considerations and Responsible Use}
\label{app:ethics}

CurveBench is designed as a controlled benchmark for evaluating visual topological reasoning. The dataset consists of hand-drawn curve images and corresponding rooted containment-tree annotations. It does not contain personal data, faces, biometric information, private text, or human-subject records.

\paragraph{Intended use.}
The intended use of CurveBench is to evaluate and improve models on visual topological reasoning, region-containment understanding, diagram interpretation, and verifiable multimodal reasoning. The benchmark is intended for research, diagnostic evaluation, and reproducible comparison of vision-language models.

\paragraph{Out-of-scope use.}
CurveBench should not be used as the sole basis for deploying models in safety-critical settings such as medical imaging, autonomous navigation, legal decision-making, military analysis, or surveillance. Performance on CurveBench measures a narrow form of region-containment reasoning and should not be interpreted as evidence of general spatial intelligence or deployment readiness.

\paragraph{Potential dual use.}
Methods that improve topological reasoning and map vectorization may have beneficial applications in document understanding, education, scientific diagram analysis, cartography, and geographic information systems. However, similar capabilities could also be adapted for sensitive geospatial analysis, surveillance, military mapping, or automated interpretation of restricted imagery. We therefore recommend that downstream users evaluate application-specific risks before deployment.

\paragraph{Dataset bias and representativeness.}
The hand-authored examples may reflect the drawing conventions and visual styles of the annotators. As a result, the dataset may underrepresent other diagrammatic conventions, cultural drawing styles, map-design conventions, or noisy real-world visual inputs. The topographical and maze-like subsets increase visual diversity, but CurveBench remains a controlled benchmark rather than a representative sample of all possible contour, map, or diagram images.

\paragraph{Limitations.}
CurveBench isolates rooted region-containment topology. It does not evaluate all aspects of spatial reasoning, such as metric geometry, object affordances, 3D structure, temporal dynamics, or real-world map semantics. The dataset size is also modest relative to large-scale vision-language datasets. Its primary purpose is diagnostic evaluation rather than large-scale pretraining.

\end{document}